\documentclass[letterpaper]{article}
\usepackage[preprint]{aaai2027}
\usepackage[hyphens]{url} % fixed
\usepackage{graphicx} % fixed
\usepackage{natbib} % fixed, without options
\usepackage{caption} % fixed, without options
\usepackage{amsmath,amssymb,booktabs,array,multirow}
\usepackage{xcolor}
\usepackage{pgfplots}
\pgfplotsset{compat=1.16}
\providecommand{\href}[2]{#2} % aaai2027.bst emits \href; AAAI forbids hyperref
\newcommand{\ci}[2]{\ensuremath{\,[#1,\,#2]}}
\newcommand{\stepn}[1]{\tikz[baseline=(s.base)]\node[circle,fill=black!88,text=white,inner sep=0.5pt,minimum size=1.05em,font=\sffamily\bfseries\scriptsize](s){#1};}
\newcommand{\figdesign}{%
\begin{figure*}[t]
\centering
\includegraphics[width=\textwidth]{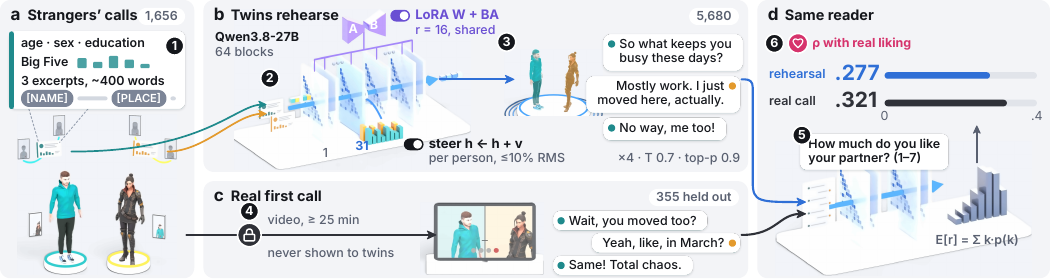}
\caption{\textbf{Study design.} \textbf{a},~\stepn{1}~Each stranger's other CANDOR calls (1,656 calls in all) give a persona card, names and places masked. \textbf{b},~\stepn{2}~The card prompts a twin, one open 27B model; the $2\times2$ arms switch on a LoRA shared by all people (rank 16) and/or a person-specific steering vector added at block 31 (at most 10\% of the hidden-state RMS). \stepn{3}~Twins meet, four times per pair. \textbf{c},~\stepn{4}~The pair's real first call is held out (355 participant-disjoint pairs). \textbf{d},~\stepn{5}~The same reader, the base model, answers ``How much do you like your partner?'' after either call; its probabilities $p(k)$ over 1--7 give an expected rating. \stepn{6}~Spearman $\rho$ with the two people's own liking. People and dialogue are illustrative; no CANDOR or twin text is shown.}
\label{fig:design}
\end{figure*}}
\newcommand{\figsignal}{%
\begin{figure*}[t]
\centering
\includegraphics[width=\textwidth]{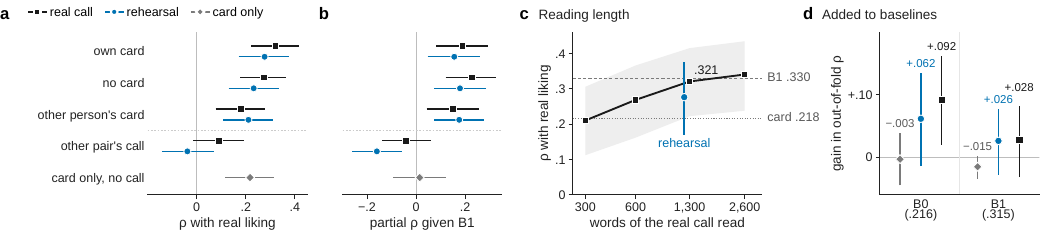}
\caption{\textbf{The rehearsal's signal travels through the conversation.} \textbf{a},~$\rho$ with real liking when the same reader reads the real call (squares) or the rehearsal (circles) with the person's own card, no card, another person's card, or another pair's call; diamond: card alone. \textbf{b},~The same as partial $\rho$ given B1. \textbf{c},~$\rho$ against the words of the real call read; the rehearsal (about 1,200 words) lies near the curve. \textbf{d},~Out-of-fold gain when each reading is added to B0 or B1 (their own $\rho$ in brackets). 355 pairs; 95\% CIs by pair bootstrap, Fisher $z$ for \emph{other person's card}.}
\label{fig:signal}
\end{figure*}}
\newcommand{\figpersonal}{%
\begin{figure*}[t]
\centering
\includegraphics[width=\textwidth]{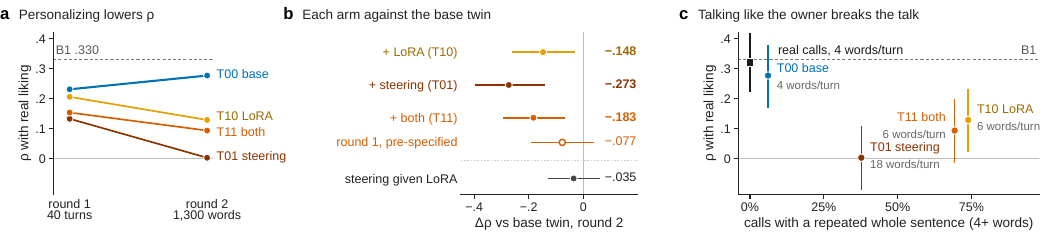}
\caption{\textbf{Making twins sound like their owners does not help.} \textbf{a},~$\rho$ of the four arms in the pre-specified round~1 (40 turns) and the exploratory round~2 (1,300 words, base-model reader); dashed: B1. \textbf{b},~Round-2 difference from the base twin (bold: 95\% CI excludes 0), the pre-specified round-1 contrast (open) and the effect of steering on top of the LoRA. \textbf{c},~Share of round-2 calls (1,420 per arm) in which a speaker repeats a whole sentence of four or more words, with median words per turn, against $\rho$. Lines: 95\% CIs (pair bootstrap for T00 and real calls, Fisher $z$ otherwise).}
\label{fig:personal}
\end{figure*}}
\newcommand{\figjudgment}{%
\begin{figure*}[t]
\centering
\includegraphics[width=\textwidth]{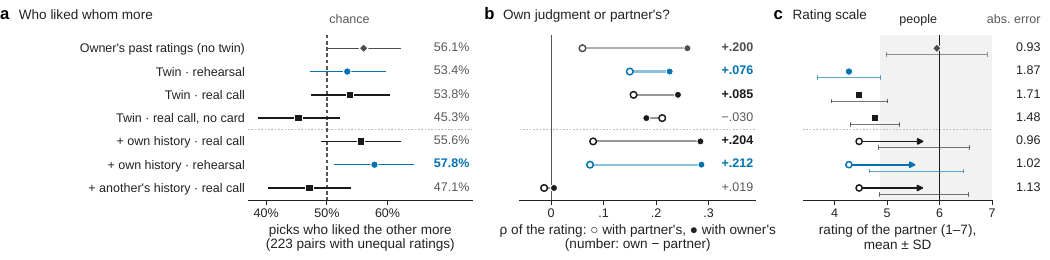}
\caption{\textbf{A persona-prompted twin holds the conversation, not yet its owner's judgment.} Rows: sources of a person's rating of their partner (710 ratings; people's mean 6.0, SD 1.13; partners' ratings correlate at 0.245); below the dotted line the twin is told a rating history. \textbf{a},~Accuracy at picking who liked the other more (223 pairs with unequal ratings); bold: 95\% CI excludes chance. \textbf{b},~Correlation with the partner's rating (open) and the owner's (filled); numbers: the difference, bold where its CI excludes 0. \textbf{c},~Mean rating $\pm$ SD; arrows run from the twin without history (open) to the twin told a history; band: people's mean $\pm$ SD; right: mean absolute error against the owner's rating. 95\% CIs: 2,000 bootstrap draws over pairs.}
\label{fig:judgment}
\end{figure*}}
\newcommand{\figpitfalls}{%
\begin{figure*}[t]
\centering
\includegraphics[width=\textwidth]{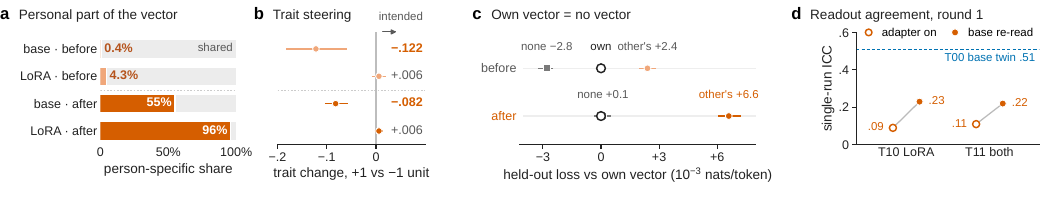}
\caption{\textbf{Diagnostics that training loss and routine output checks did not show.} Light: steering vectors before retraining; dark: after. \textbf{a},~Person-specific share of the vector's energy. \textbf{b},~IPIP-50 score of the targeted trait steered at $+1$ minus $-1$ unit (five-trait mean). \textbf{c},~Held-out loss on each person's own speech with the LoRA (710 people), relative to one's own vector (open). \textbf{d},~Single-run ICC of the liking readout on round-1 calls with the adapter on (open) and after a base-model re-read (filled); dashed: base twin. Lines: 95\% CIs, 5,000 bootstrap draws over items (b) or people (c).}
\label{fig:pitfalls}
\end{figure*}}
\newcommand{\figregimes}{%
\begin{figure}[!tb]
\centering
\includegraphics[width=\columnwidth]{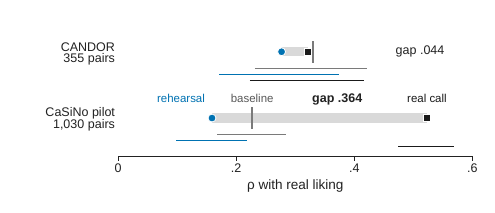}
\caption{\textbf{The same-reader ceiling locates the bottleneck.} $\rho$ of the same reader on the twins' rehearsal (circle) and on the real interaction (square); grey tick: the non-LLM baseline (CANDOR: B1; CaSiNo: pre-negotiation ridge). In CANDOR the rehearsal nearly reaches the ceiling, so the limit is the judgment; in a pre-specified CaSiNo pilot it falls far short, so the simulation loses the signal. Thin lines: 95\% CIs (pair bootstrap for the CANDOR reader, Fisher $z$ otherwise).}
\label{fig:regimes}
\end{figure}}
\newcommand{\figappraisal}{%
\begin{figure*}[t]
\centering
\includegraphics[width=\textwidth]{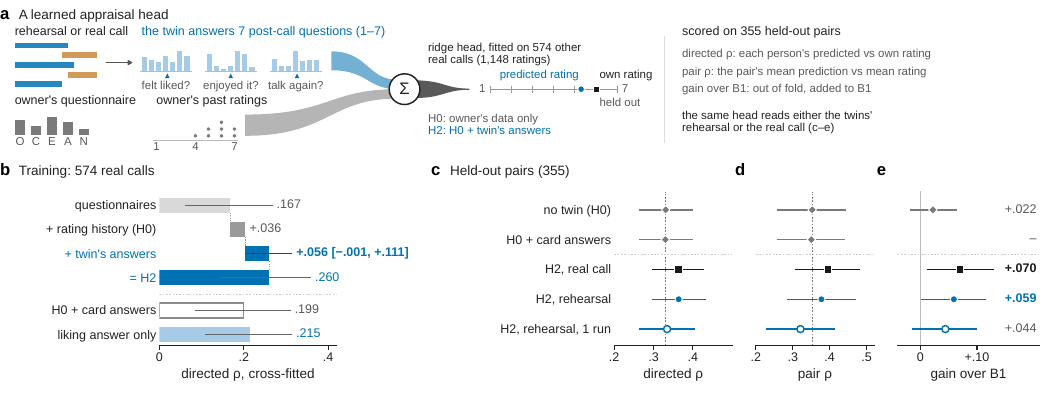}
\caption{\textbf{A learned appraisal head turns the twin's post-call answers into usable ratings} (exploratory). \textbf{a},~A ridge head fitted only on 574 non-held-out real calls (1,148 ratings) combines the twin's expected answers to seven post-call questions with the owner's questionnaire (Big Five: O, C, E, A, N) and rating history (H2), or uses the owner's data alone (H0); glyphs are illustrative. \textbf{b},~Directed $\rho$, cross-fitted over the training calls; blue step: the pre-specified contrast H2 $-$ H0, whose interval just reaches zero (none of 200 label permutations reached it). \textbf{c--e},~Held-out pairs: directed $\rho$, pair $\rho$ and out-of-fold gain over B1 (0.315); dashed: H0; open circle: one rehearsal instead of four. 95\% CIs: bootstrap over people (b) or pairs (c--e).}
\label{fig:appraisal}
\end{figure*}}

\title{Can Your AI Twin Meet People for You?\\It Can Hold a Conversation. But What Does It Need to Judge Like You?}
\author{
\normalfont\fontsize{10}{11.5}\selectfont
\makebox[0.96\textwidth][s]{\mbox{Wanghao Ye\textsuperscript{1}}\hfill \mbox{Sihan Chen\textsuperscript{2}}\hfill \mbox{Bowei Tian\textsuperscript{1}}\hfill \mbox{Yiting Wang\textsuperscript{1}}\hfill \mbox{Shwai He\textsuperscript{1}}\hfill \mbox{Guoheng Sun\textsuperscript{1}}\hfill \mbox{Ziyi Wang\textsuperscript{3}}\hfill \mbox{Ziyao Wang\textsuperscript{1}}\hfill \mbox{Yexiao He\textsuperscript{1}}}\\
\makebox[0.96\textwidth][s]{\mbox{Zheyu Shen\textsuperscript{1}}\hfill \mbox{Meng Liu\textsuperscript{1}}\hfill \mbox{Yuning Zhang\textsuperscript{1}}\hfill \mbox{Meng Feng\textsuperscript{1}}\hfill \mbox{Yanhong Qian\textsuperscript{1}}\hfill \mbox{Yang Wang\textsuperscript{4}}\hfill \mbox{Siyuan Peng\textsuperscript{1}}}\\
\makebox[0.96\textwidth][s]{\mbox{Yilong Dai\textsuperscript{5}}\hfill \mbox{Zhenle Duan\textsuperscript{1}}\hfill \mbox{Joshua Liu\textsuperscript{2}}\hfill \mbox{Lang Xiong\textsuperscript{6}}\hfill \mbox{Aarosh Das\textsuperscript{1}}\hfill \mbox{Zhaoyi Liu\textsuperscript{1}}\hfill \mbox{Daeun Jung\textsuperscript{1}}\hfill \mbox{Aditya Sarkar\textsuperscript{1}}}\\
\makebox[0.96\textwidth][s]{\mbox{Parastoo Pilevar\textsuperscript{1}}\hfill \mbox{Xiaochuan Gong\textsuperscript{1}}\hfill \mbox{Yancheng Zhu\textsuperscript{1}}\hfill \mbox{Yingying Fan\textsuperscript{1}}\hfill \mbox{Qiong Wang\textsuperscript{1}}\hfill \mbox{Hanzhang Qin\textsuperscript{7,*}}\hfill \mbox{Ang Li\textsuperscript{1,*}}}
}
\affiliations{
\normalfont\fontsize{8.5}{10}\selectfont
\mbox{\textsuperscript{1}University of Maryland, College Park}\hspace{0.65em}\mbox{\textsuperscript{2}Carnegie Mellon University}\hspace{0.65em}\mbox{\textsuperscript{3}Texas A\&M University, College Station}\\
\mbox{\textsuperscript{4}Northeastern University}\hspace{0.65em}\mbox{\textsuperscript{5}The University of Alabama}\hspace{0.65em}\mbox{\textsuperscript{6}Stanford University}\hspace{0.65em}\mbox{\textsuperscript{7}National University of Singapore}\\[-1pt]
\textsuperscript{*}Co-corresponding authors: Hanzhang Qin and Ang Li.
}

\begin{document}
\maketitle

\begin{abstract}
Should your AI twin go meet people on your behalf? We built LLM twins of 710 real people from their other conversations in the CANDOR corpus, sent them to meet each other in 5,680 simulated first conversations, and checked every rehearsal against what the real people said after meeting for real (355 held-out, participant-disjoint pairs, against baselines that include each person's rating history). The rehearsals work: a twin's conversation predicts real liking ($\rho=0.277$), carries information that questionnaires and rating history do not have (partial $\rho=0.16$\ci{0.05}{0.26}), and recovers most of what the same reader gets from the real conversation (0.321). Swap controls show that this signal travels through the conversation, not through the persona card. What, then, does a twin need to judge like you? Not your voice: making twins talk more like their owners, with person-specific activation steering and a LoRA trained on their own speech, did not help in a pre-specified test and lowered accuracy in an exploratory second round ($-0.183$\ci{-0.294}{-0.066}). Nor simply your stated ratings: a persona-prompted twin that holds the conversation does not yet carry its owner's judgment, and when told its owner's past ratings it anchors on them instead of weighing them. Learning helps: a small appraisal head trained on other real conversations, which combines the twin's post-call answers with its owner's rating history, lifts rehearsal-based prediction above the history baseline out of fold ($+0.059$\ci{0.001}{0.115}, against $+0.022$ without the twin's answers; pre-specified primary contrast $+0.056$\ci{-0.001}{0.111}). We propose an attribution protocol for twin-based social simulation and identify learned appraisal as the component twins need next.
\end{abstract}

\section{Introduction}

Digital twins built from large language models are moving from answering surveys to acting in social settings. Models conditioned on people's backgrounds reproduce survey subgroups and classic experiments \citep{argyle2023outofone,aher2023turing}, and agents grounded in a person's interviews, answers or profile reproduce much of their owner's survey behavior \citep{park2024agents,toubia2025twin,chen2025personatwin}. The next step is to let them act for us. In social matching, the twins of two candidates meet first, and the pairs whose twins hit it off are recommended to meet in person: twin architectures for social pairing report agreement with real speed-dating decisions \citep{ye2025cognipair}, and deployed platforms let twins converse on their users' behalf to estimate compatibility \citep{ye2026cognibit}. What has not been established is where in the pipeline that signal lives: in the simulated conversation, or in the twin's judgment of it. Twins are usually validated on whether they answer questionnaires like their owners or sound human, and neither check separates these two steps. Separating them is the attribution problem of LLM social simulation.

We run that test. In the CANDOR corpus \citep{reece2023candor}, 1,456 people had 1,656 unscripted video calls of at least 25 minutes with strangers and afterwards rated how much they liked their partner. Because many people took part in several calls, we build each person's twin only from their \emph{other} calls, let the twins of a held-out pair meet, and compare each twin's liking with what the two real people answered. Every held-out comparison uses 355 pairs that share no participants and is made against non-LLM baselines built from questionnaires and each person's rating history. Each experiment gives the twin one kind of information: the encounter, its owner's persona card, speaking style (pre-specified) or past ratings.

\figdesign

Figure~\ref{fig:design} shows the design; we make three points:
\begin{itemize}
\item \textbf{Twin rehearsals carry real social signal.} A base twin's simulated conversation predicts real liking beyond questionnaires and rating history and recovers most of what the same reader extracts from the real call; swap controls place this signal in the conversation (\S\ref{sec:signal}).
\item \textbf{Sounding like the owner is not the lever.} Person-specific steering and a LoRA trained on the owner's speech did not help in a pre-specified $2\times2$ test and lowered accuracy in an exploratory second round (\S\ref{sec:personalization}). We document six failure modes of persona steering and twin readouts that standard checks miss (\S\ref{sec:pitfalls}).
\item \textbf{Judgment is the next step.} At the person level, persona-prompted twins do not yet reliably tell who liked whom more, and they echo a stated rating history rather than weighing it; a first learned appraisal head already helps at the pair level (\S\ref{sec:verdict}). We propose an attribution protocol and name learned appraisal the next design target (\S\ref{sec:discussion}).
\end{itemize}

\section{Related Work}

\paragraph{LLM social simulation.} LLM agents populate simulated towns \citep{park2023generative}, goal-driven encounters \citep{zhou2024sotopia} and societies of over 10,000 agents \citep{piao2025agentsociety}, where conventions and collective biases emerge \citep{ashery2025conventions}, prosocial behavior responds to policy \citep{zhou2026prosim}, and parallel counterfactual runs compare moderation strategies with all else equal \citep{fidone2026cosmos}. The method's promise rests on verifiable comparison with human data \citep{anthis2025position}: the best models reach only modest fidelity to human response distributions \citep{hu2026simbench}, validity comes from calibration against human data rather than heuristic checks \citep{hullman2026validating}, and persona studies often leave their target population underspecified \citep{batzner2025whose}.

\paragraph{Digital twins of individuals.} Simulations grounded in two-hour interviews and surveys match individuals' General Social Survey answers 72.8\% of the time, against 79.5\% for the people themselves two weeks later \citep{park2024agents}; twin panels with 500 answers per person are public \citep{toubia2025twin}. Across 164 outcomes, twin and human answers correlate at 0.20 on average, with insufficient individuation \citep{peng2025megastudy}; persona variables explain under 10\% of annotation variance \citep{hu2024personaeffect}, LLM stand-ins flatten identity groups \citep{wang2025misportray}, models cannot play personas that perform worse than themselves \citep{kumar2025reversed}, more than half of value-conditioned personas fail to express their profile from the outset \citep{atif2026drift}, and personas drift over multi-turn dialogue unless trained for consistency with multi-turn RL \citep{abdulhai2025consistent}. We add an interactive, dyadic outcome, where the twin's decision depends on a partner it has never met.

\paragraph{Steering and personalization.} Persona or trait directions can be read, added or trained in activation space \citep{rimsky2024caa,chen2025persona,cao2024bipo}, and adapters \citep{hu2022lora} can imitate a person's style. They are usually evaluated with Big-Five questionnaires \citep{goldberg1992markers}, yet strong steering pushes free text into repetition \citep{braun2026beyond}, role directions need per-role steering strengths \citep{song2026rolesteering}, the advantage of feature- and probe-based steering over prompting depends on the prompting strategy \citep{fan2026steering}, and personality LoRAs role-play no better than prompting \citep{krabbe2026personality}; we evaluate both on social outcomes.

\paragraph{Predicting attraction and judging conversations.} In speed dating, the largest share of the variance in desire is specific to the pair, and self-reports collected before the dates could not predict it \citep{joel2017desire}; romantic evaluations are more unique to the pair than consensual \citep{eastwick2014relational}, so matching before people meet is unlikely to explain more than a small part of long-term outcomes \citep{finkel2012online}, and people underestimate how much a new partner likes them \citep{boothby2018liking}. LLMs reading speed-dating transcripts predict matching only modestly \citep{matz2024attraction}, LLM judges of conversational success on CANDOR correlate only moderately with participants' ratings \citep{zufle2026judges}, and multi-question LLM judgments combined by a small calibrated network with judge-specific parameters agree better with individual raters \citep{hashemi2024rubric}, motivating our appraisal head.

\figsignal

\section{Setup}
\label{sec:setup}

\paragraph{Data and split.} Figure~\ref{fig:design} summarizes the design; the numbered badges below refer to its steps. The outcome is each person's answer to ``How much do you like your conversation partner?'' (1--7) after the call; the pair-level target is the mean of the two answers. Among calls in which both people answered and both had at least one other call, we held out 355 calls by maximum-cardinality matching so that no person appears in two held-out calls (710 people). Held-out calls are excluded from every training and conditioning input; the pair's real first call is used only for scoring (Fig.~\ref{fig:design}c, \stepn{4}).

\paragraph{Twins.} Each twin is an open-weight 27B instruction model (Qwen3.8-27B; Fig.~\ref{fig:design}b, \stepn{2}), run locally because the data agreement rules out external services. Its system prompt holds a persona card built only from the person's other calls (Fig.~\ref{fig:design}a, \stepn{1}): age, sex, education, Big Five self-ratings, and three excerpts of about 400 words each from those calls, with names, organizations and places masked. The twin is told to talk the way it talks in the excerpts. Two twins alternate turns, each seeing only its own card (temperature 0.7, top-$p$ 0.9). Each held-out pair is rehearsed four times (\stepn{3}).

\paragraph{Readout.} After the call each twin privately answers the same liking question (Fig.~\ref{fig:design}d, \stepn{5}). We use the expectation of the model's probabilities over the digits 1--7, average the two sides and the four runs, and correlate with the real pair mean (Spearman $\rho$, \stepn{6}).

\paragraph{Metrics.} The \emph{reader} is the base model with the owner's persona card; in round~2 it rates rehearsals and real calls alike. Pair-level $\rho$ compares the mean of the two ratings with the real pair mean; the \emph{directed} correlation compares each of 710 people's predicted rating with their own rating of the partner. Partial $\rho$ given B1 removes B1's prediction from both sides, and an \emph{out-of-fold gain} is the change in $\rho$ when a reading is added to a baseline, scored on folds not used for fitting. Intervals: 95\%, paired bootstrap over held-out pairs (5,000 draws; 2,000 for person-level analyses).

\paragraph{Personalization arms.} A $2\times2$ design (Fig.~\ref{fig:design}b): T00 is the base twin; T10 adds one LoRA adapter (rank 16) shared by all people, trained with the loss on each person's own turns in their excerpts; T01 adds a person-specific steering vector in the residual stream, combining a per-person component with a learned map of the person's Big Five scores and trained on the person's own excerpts; T11 adds both.

\paragraph{Comparators.} B0 is a ridge regression on both people's questionnaires; B1 adds each person's rating history from their other calls (mean liking given, received and enjoyment), information the twins never see. Both are fitted on 574 non-held-out calls. C00 asks the base model to read both persona cards and predict the liking directly, without a conversation.

\paragraph{Two rounds.} Round 1 is pre-specified: before any outcome was read we froze the split, materials, arms, baselines, the primary contrast $\rho(\mathrm{T11})-\rho(\mathrm{T00})$ and the analysis (paired bootstrap, 95\% percentile intervals), and sealed predictions before loading outcomes. Round 1 calls ran 40 turns and each arm read its own ratings. Round 2 is exploratory and incorporates the diagnostics of \S\ref{sec:pitfalls}: steering vectors retrained so that the person-specific part carries most of their energy (Fig.~\ref{fig:pitfalls}a), calls run to 1,300 words or 120 turns, and every rating read by the plain base model. All 5,680 rehearsals were regenerated. Round-2 analyses use outcomes already seen and are not corrected for multiple comparisons.

\figpersonal

\section{Results}

\subsection{Twin rehearsals carry the signal}
\label{sec:signal}

The base twin's rehearsal predicts real liking at $\rho=0.277$ (Table~\ref{tab:main}), above the questionnaire model B0 (0.242). The same reader applied to the \emph{real} held-out calls, cut by the same rule, gives a ceiling of 0.321, and the rehearsal is not significantly below it ($+0.044$\ci{-0.051}{0.142}). Beyond the strongest baseline B1 (0.330), the rehearsal carries independent information: its partial correlation given B1 is 0.156\ci{0.047}{0.262}, against 0.189\ci{0.081}{0.293} for the real call, and adding either to B1 raises the out-of-fold correlation from 0.315 to about 0.34 ($+0.026$\ci{-0.028}{0.077} and $+0.028$\ci{-0.032}{0.083}), a smaller and non-significant gain because it also pays for fitting the combination. Conversation adds most for newcomers without rating history (Fig.~\ref{fig:signal}d): over the questionnaire model (out-of-fold 0.216), the real call adds $+0.092$\ci{0.021}{0.163} and the rehearsal $+0.062$\ci{-0.014}{0.134}. Reading more of the real call helps up to about 1,300 words (Fig.~\ref{fig:signal}c), so the 1,300-word cut keeps most of what the reader can use.

\begin{table}[t]
\centering\small
\setlength{\tabcolsep}{4pt}
\begin{tabular}{@{}l@{\hspace{5pt}}lcccr@{}}
\toprule
 & & Round 1 & \multicolumn{3}{c}{Round 2}\\
\cmidrule(lr){3-3}\cmidrule(l){4-6}
 & Predictor & $\rho$ & $\rho$ & ICC$_1$ & Repeat\\
\midrule
B0  & Questionnaires (ridge)    & .242 & .242 & -- & --\\
B1  & B0 + rating history       & \textbf{.330} & \textbf{.330} & -- & --\\
C00 & LLM reads both cards      & .181 & --   & -- & --\\
    & Reader, own card, no call & --   & .218 & -- & --\\
\midrule
T00 & Base twin                 & .231 & .277 & .41 & 6\%\\
T10 & + shared LoRA             & .206 & .129 & .21 & 74\%\\
T01 & + person steering         & .133 & .004 & .33 & 38\%\\
T11 & + both                    & .154 & .094 & .26 & 69\%\\
\midrule
    & Real call, same reader    & --   & .321 & -- & 0\%\\
\bottomrule
\end{tabular}
\caption{Spearman $\rho$ with the real pair-mean liking (355 held-out pairs). ICC$_1$: agreement between single rehearsals. Repeat: share of calls containing a repeated whole sentence of four or more words.}
\label{tab:main}
\end{table}

Figure~\ref{fig:signal}a,b shows where the signal travels. Swapping in another pair's conversation erases it (0.090 real, $-0.038$ rehearsal), while removing the reader's persona card, or giving it someone else's, leaves most of it. Reading the card alone reaches 0.218 but adds nothing beyond B1 (partial 0.014). What a twin contributes beyond questionnaires and history therefore arrives through the conversation. Pretraining recall of CANDOR cannot explain it: rehearsals are newly generated, held-out calls never enter twin construction, and the swaps show that the signal needs this pair's conversation. Rehearsals are generated from both partners' cards, so part of this is persona information expressed in dialogue; the real-call rows show the same reader drawing comparable independent signal from what people actually said.

\subsection{Sounding like the owner is not the lever}
\label{sec:personalization}

\figpitfalls

The pre-specified question was whether personalizing how twins talk makes rehearsals more predictive. It does not (Fig.~\ref{fig:personal}a): between rounds only the base twin gains (0.231 to 0.277), while every personalized arm falls. In round 1 the primary contrast T11$-$T00 was $-0.077$\ci{-0.192}{0.039}, and B1 stayed ahead of every twin arm (for T11 by $0.176$\ci{0.055}{0.297}). With round 2's base-model reader (single-run ICC 0.21--0.41, against 0.09--0.11 for LoRA arms read through the adapter in round~1) the losses are significant: T11$-$T00 $=-0.183$\ci{-0.294}{-0.066}, LoRA alone $-0.148$\ci{-0.263}{-0.030}, steering alone $-0.273$\ci{-0.398}{-0.142}; adding steering on top of the LoRA changes little ($-0.035$\ci{-0.130}{0.063}; Fig.~\ref{fig:personal}b). The rehearsals show why. Twins carrying the LoRA repeat whole sentences of four or more words in about 70\% of calls, which real CANDOR speakers never do; the base twin does so in 6\% (Fig.~\ref{fig:personal}c). Steering alone repeats less (38\%) but stretches turns to a median of 18 words, against 4 in real calls. Because the signal travels through the conversation, personalization that makes twins sound like their owners by degrading what they say costs more than it adds.

\subsection{What standard checks miss}
\label{sec:pitfalls}

Six failure modes behind these results do not show in training loss or routine output checks; each surfaces only under a targeted diagnostic (Table~\ref{tab:pitfalls}, Fig.~\ref{fig:pitfalls}). Retraining made the vectors personal (Fig.~\ref{fig:pitfalls}a), yet steering a trait up still lowers its questionnaire score on the base model (b), one's own vector only ties no vector (c), and a base-model re-read improves but does not restore the readout's run-to-run agreement (d).

\begin{table}[!tb]
\centering\footnotesize
\setlength{\tabcolsep}{3pt}
\begin{tabular}{@{}>{\raggedright\arraybackslash}p{2.0cm}>{\raggedright\arraybackslash}p{3.85cm}>{\raggedright\arraybackslash}p{2.05cm}@{}}
\toprule
Failure mode & What we observed & Exposed by\\
\midrule
Personal vector mostly shared & Per-person part held 0.4\% of the energy; 79\% of steered calls looped & Energy split (Fig.~\ref{fig:pitfalls}a)\\
\addlinespace[3pt]
Trait direction is reversed & Steering $+1$ vs.\ $-1$ trait unit changed the IPIP-50 score by $-0.08$\ci{-0.10}{-0.06} & Steered IPIP-50 (Fig.~\ref{fig:pitfalls}b)\\
\addlinespace[3pt]
Own vector ties no vector & Held-out loss vs.\ own vector ($10^{-3}$ nats/token): other's $+6.6$, none $+0.1$ & Own / other / none loss (Fig.~\ref{fig:pitfalls}c)\\
\addlinespace[3pt]
Adapter flattens the ratings & Round 1: single-run ICC 0.09--0.11 with the LoRA, 0.51 without & Run-to-run ICC (Fig.~\ref{fig:pitfalls}d)\\
\addlinespace[3pt]
Turn caps keep only openings & 40 turns cover only the start of a 25-minute call & Cap on words, not turns\\
\addlinespace[3pt]
Adapter leaks into the readout & Round-1 LoRA arms were rated through the adapter & Re-read by the base model (Fig.~\ref{fig:pitfalls}d)\\
\bottomrule
\end{tabular}
\caption{Failure modes of persona steering and twin readouts. Round 2 fixes rows 1 and 4--6; rows 2--3 persist after retraining and bound what steering can deliver.}
\label{tab:pitfalls}
\end{table}

\subsection{A persona-prompted twin holds the conversation, not yet its owner's judgment}
\label{sec:verdict}

The conversation is not where the signal is lost: even on real transcripts, the reader's pair-level accuracy only matches B1 ($-0.009$), so the limiting step is the post-call judgment. A twin stands in for one person, so Figure~\ref{fig:judgment} scores each twin's rating against its owner's rating of that partner. Twin ratings correlate with their owners' at 0.23--0.24 and favor the owner over the partner ($+0.08$\ci{0.01}{0.15}; Fig.~\ref{fig:judgment}b), so they do carry person-specific information. But they pick who liked the other more in only 53--54\% of pairs (Fig.~\ref{fig:judgment}a), with intervals that include chance, while the owner's own past ratings reach 56\%. The person-specific part comes from the card: without it the twin no longer favors its owner, and its accuracy falls to chance level. Twins are also harsher than people, rating partners 4.3--4.5 where people say 6.0 (Fig.~\ref{fig:judgment}c).

Telling a twin its owner's habits fixes the scale but not the judgment. One sentence with the person's mean past rating moves the twin's mean to 5.71 (Fig.~\ref{fig:judgment}c), nearly halves its error and lifts the directed correlation to 0.284 ($+0.04$\ci{-0.05}{0.14} over no history). The twin's own rating complements the history (partial correlation given history 0.197), but once told the history the twin anchors on it: with another person's history the correlation drops to 0.005 (own minus other's: $+0.279$\ci{0.18}{0.37}), the partial correlation falls to 0.137, and a plain statistical combination of the history-free rating with the history does at least as well (0.301). The same holds after rehearsals, where the twin told its owner's history reaches 0.286 and rates as it does after the real call (directed difference $-0.002$\ci{-0.03}{0.02}). Told its owner's history, the rehearsal twin picks who liked whom more in 57.8\% of pairs, the only twin row whose interval excludes chance and about what the history alone gives (56.1\%; Fig.~\ref{fig:judgment}a); adding how past partners rated the owner changes nothing further (directed $-0.009$\ci{-0.02}{0.00}).

\figjudgment

\paragraph{Learning to judge.} Can the twin's own post-call answers become informative once it learns how people turn such impressions into ratings? On 574 other real calls (none held out), we fitted a small ridge head that combines the twin's answers to seven post-call questions (such as whether it felt liked, enjoyed the call or would talk again) with the owner's questionnaire and rating history, and compared it with the same head without the twin's answers; we call these heads H2 and H0 (Fig.~\ref{fig:appraisal}a). Cross-fitted over the training calls, the twin's answers raise the directed correlation from 0.204 to 0.260 (Fig.~\ref{fig:appraisal}b) ($+0.056$\ci{-0.001}{0.111}, the pre-specified primary contrast, whose interval just reaches zero; none of 200 label permutations of the whole pipeline reached this gain), while the same questions answered from the persona card alone add nothing ($-0.005$), so the gain comes from the conversation. Trained on real calls only, the head reaches a pair-level correlation of 0.394 on the held-out real calls and 0.378 on the rehearsals (Fig.~\ref{fig:appraisal}c--e), and improves on B1 out of fold ($+0.070$\ci{0.011}{0.130} and $+0.059$\ci{0.001}{0.115}, averaging four rehearsals); its ratings sit on people's scale (mean 5.89 against 6.0). The gain carries over from real calls to rehearsals without measurable loss (difference $-0.001$\ci{-0.052}{0.050}), while who liked whom more still comes from the owner's history (about 61\% with or without the twin's answers). Part of the lift comes from learning to weigh the history: H0 alone already gains $+0.022$\ci{-0.017}{0.064} over B1, and on held-out pairs the twin's answers add $+0.033$\ci{-0.004}{0.072} in directed $\rho$ after rehearsals (Fig.~\ref{fig:appraisal}c,e), so the pre-specified contrast on the training calls is our main evidence that the answers help.

\section{Discussion}
\label{sec:discussion}

\begin{table}[t]
\centering\footnotesize
\setlength{\tabcolsep}{3pt}
\begin{tabular}{@{}>{\raggedright\arraybackslash}p{2.2cm}>{\raggedright\arraybackslash}p{2.05cm}>{\raggedright\arraybackslash}p{3.65cm}@{}}
\toprule
Given to the twin & Effect & Evidence\\
\midrule
The encounter & Adds signal beyond B1 & Partial $\rho$ .156\ci{.05}{.26}; another pair's call: $\rho$ $-$.038\\
\addlinespace[3pt]
Persona card$^\dagger$ & Ties the rating to its owner & Card alone: partial .014; own $-$ partner $+$.085, without card $-$.030\\
\addlinespace[3pt]
Speaking style of the owner & Lower accuracy & LoRA $+$ steering: $-$.183\ci{-.29}{-.07}\\
\addlinespace[3pt]
Past ratings in the prompt$^\dagger$ & Sets the scale; twin anchors & Mean 4.47 $\to$ 5.71; another's history: $\rho$ .005\\
\addlinespace[3pt]
How partners rated the owner$^\dagger$ & No further gain & $-$.009\ci{-.02}{.00}\\
\addlinespace[3pt]
Learned head: past ratings $+$ post-call answers & Lifts prediction above B1 & $+$.059\ci{.001}{.115} out of fold; without the twin's answers $+$.022\\
\bottomrule
\end{tabular}
\caption{What a twin needs to judge like you (355 held-out pairs; $^\dagger$read on real calls, others on rehearsals). Pre-specified: the speaking-style test (round~1, no gain) and the head's primary contrast (Fig.~\ref{fig:appraisal}); values shown are exploratory.}
\label{tab:recipe}
\end{table}

\paragraph{What a twin needs to judge like you.} Table~\ref{tab:recipe} collects the answer. The owner's past judgments carry what the twin's judgment lacks, and stating them in the prompt makes the twin anchor on them rather than integrate them; imitating the owner's voice did not help. Appraisal should therefore be a component of the twin in its own right, learned from the person's past evaluations and validated person by person. Agent architectures with explicit memory \citep{park2023generative}, including twins whose preferences evolve across encounters \citep{ye2025cognipair}, are natural places to integrate these evaluations, and the protocol below measures the gain in the owner's own judgment. A first learned appraisal head already lifts rehearsal-based prediction above the history baseline (\S\ref{sec:verdict}); learning each owner's taste, rather than their leniency, will need more than one to three past ratings per person.

\figappraisal

\paragraph{An attribution protocol.} For any twin used to predict social outcomes we recommend reporting: (i) a baseline with the same pre-interaction information, including rating history; (ii) the same reader applied to real interactions, as a ceiling that locates any loss in the simulation or in the judgment; (iii) swap controls for the persona and the interaction, which separate signal from prompting artifacts and recall; (iv) person-level, directed agreement and who-liked-more accuracy rather than pair means only; (v) the own-versus-other-person control for any personalization; and (vi) verbatim repetition against real dialogue.

\figregimes

\paragraph{When the rehearsal is the bottleneck.} The same protocol identifies settings where the simulation, not the judgment, loses the signal. In a pre-specified pilot on post-negotiation liking (CaSiNo, \citealt{chawla2021casino}; 1,030 pairs), the same reader reached 0.523 on real negotiations but 0.159 on twin-to-twin negotiations, below a ridge baseline (0.227). Comparing a rehearsal with its same-reader ceiling separates the two cases (Fig.~\ref{fig:regimes}).

\paragraph{Scope.} CANDOR measures liking after one conversation between strangers rather than dating outcomes, on a single item near its ceiling (mean 6.0 of 7), and we read text only; these bound what any text reader can reach. Each held-out person meets one partner and has one to three past ratings, which bounds person-level models. Our twins are persona-prompted base models without dedicated cognitive modules, so the judgment results describe this simplest design. We use one open-weight model, as the data agreement excludes external services. Our arms test two common ways to personalize style, not owner-likeness itself: the retrained vector fits held-out speech no better than none (Fig.~\ref{fig:pitfalls}c).

\section{Conclusion}
AI twins can already hold the conversation: their rehearsals predict real liking beyond questionnaires and rating history and keep most of what the same reader finds in the real call. They do not yet carry their owner's judgment: sounding like the owner lowered accuracy, and a twin told its owner's past ratings anchors on them. Learning from people's past evaluations helps, as a small appraisal head lifts rehearsal-based prediction above the history baseline out of fold. Twin-based social simulation should separate the simulation from the judgment, test each against a same-reader ceiling and swap controls, and invest in learned, person-specific appraisal rather than an imitated voice.

\section*{Broader Impact}
Twin rehearsals could let people screen many potential contacts, such as dates, cofounders, mentors or roommates, without an hour-long first meeting each; this could especially help people with little time, limited mobility or social anxiety to widen their circles, and let anyone preview how a first conversation might go before they meet. For research, rehearsals allow studies of first impressions at scale and with counterfactual control, such as replaying one encounter with a single input changed; researchers could pilot designs before recruiting anyone, or study encounters that would be impractical or unethical to stage. Our attribution protocol turns such studies into auditable evidence: a same-reader ceiling, swap controls and person-level metrics show where a twin's signal comes from, and our failure modes catch problems that training loss and routine checks miss. The protocol needs only a model's ratings and reports aggregates, so other groups can apply it to their own data without releasing anyone's conversations. For platforms, the same checks show before deployment whether to improve the simulation or the judgment, and person-level validation shows where a twin misrepresents its owner before anyone relies on it. The same results mark the risks. Twins do not yet carry their owner's judgment, so their recommendations should be hints that people confirm by meeting, not decisions made for them. Twins need personal conversation data, so consent, data minimization and the ability to inspect and delete one's twin are essential; judgment learned from past ratings can carry a person's or a platform's biases forward and should be audited person by person and across groups. Only its owner should control a twin, so that no one else can use it to probe or manipulate them.

\section*{Ethics and Data Statement}
CANDOR was used under its terms of use for non-commercial academic research. All processing ran on a secured research cluster with a locally hosted model; no transcript, persona material or generated text left the cluster or appears in this paper, and we report only aggregates.

\section*{AI Disclosure Statement}
Following the AAAI policy on AI systems: a locally hosted open-weight language model is part of the method, as the twins and readers (\S\ref{sec:setup}). The authors also used generative AI assistants for analysis code, experiment monitoring, literature search, and drafting and editing text and figures; the authors reviewed and verified all such outputs, including numbers, claims and citations, and take full responsibility for the paper.

\bibliography{refs}

\begin{thebibliography}{41}
\providecommand{\natexlab}[1]{#1}

\bibitem[{Abdulhai et~al.(2025)Abdulhai, Cheng, Clay, Althoff, Levine, and
  Jaques}]{abdulhai2025consistent}
Abdulhai, M.; Cheng, R.; Clay, D.; Althoff, T.; Levine, S.; and Jaques, N.
  2025.
\newblock Consistently Simulating Human Personas with Multi-Turn Reinforcement
  Learning.
\newblock In \emph{Advances in Neural Information Processing Systems}.

\bibitem[{Aher, Arriaga, and Kalai(2023)}]{aher2023turing}
Aher, G.; Arriaga, R.~I.; and Kalai, A.~T. 2023.
\newblock Using Large Language Models to Simulate Multiple Humans and Replicate
  Human Subject Studies.
\newblock In \emph{Proceedings of the 40th International Conference on Machine
  Learning}, volume 202 of \emph{PMLR}, 337--371.

\bibitem[{Anthis et~al.(2025)Anthis, Liu, Richardson, Kozlowski, Koch,
  Brynjolfsson, Evans, and Bernstein}]{anthis2025position}
Anthis, J.~R.; Liu, R.; Richardson, S.~M.; Kozlowski, A.~C.; Koch, B.;
  Brynjolfsson, E.; Evans, J.; and Bernstein, M.~S. 2025.
\newblock Position: {LLM} Social Simulations Are a Promising Research Method.
\newblock In \emph{Proceedings of the 42nd International Conference on Machine
  Learning}, volume 267 of \emph{PMLR}, 81005--81034.

\bibitem[{Argyle et~al.(2023)Argyle, Busby, Fulda, Gubler, Rytting, and
  Wingate}]{argyle2023outofone}
Argyle, L.~P.; Busby, E.~C.; Fulda, N.; Gubler, J.~R.; Rytting, C.; and
  Wingate, D. 2023.
\newblock Out of One, Many: Using Language Models to Simulate Human Samples.
\newblock \emph{Political Analysis}, 31(3): 337--351.

\bibitem[{Ashery, Aiello, and Baronchelli(2025)}]{ashery2025conventions}
Ashery, A.~F.; Aiello, L.~M.; and Baronchelli, A. 2025.
\newblock Emergent Social Conventions and Collective Bias in {LLM} Populations.
\newblock \emph{Science Advances}, 11(20): eadu9368.

\bibitem[{Atif, Saha, and Choudhury(2026)}]{atif2026drift}
Atif, F.; Saha, S.; and Choudhury, M. 2026.
\newblock The Failure Happens Before the Drift: The Social Dynamics of Values
  in {LLM} Agent Societies.
\newblock In \emph{Proceedings of the AAAI/ACM Conference on AI, Ethics, and
  Society (AIES)}.

\bibitem[{Batzner et~al.(2025)Batzner, Stocker, Tang, Natarajan, Chen, Schmid,
  and Kasneci}]{batzner2025whose}
Batzner, J.; Stocker, V.; Tang, B.; Natarajan, A.; Chen, Q.; Schmid, S.; and
  Kasneci, G. 2025.
\newblock Whose Personae? Synthetic Persona Experiments in {LLM} Research and
  Pathways to Transparency.
\newblock In \emph{Proceedings of the AAAI/ACM Conference on AI, Ethics, and
  Society (AIES)}, volume~8, 343--354.

\bibitem[{Boothby et~al.(2018)Boothby, Cooney, Sandstrom, and
  Clark}]{boothby2018liking}
Boothby, E.~J.; Cooney, G.; Sandstrom, G.~M.; and Clark, M.~S. 2018.
\newblock The Liking Gap in Conversations: Do People Like Us More Than We
  Think?
\newblock \emph{Psychological Science}, 29(11): 1742--1756.

\bibitem[{Braun, Eickhoff, and Bahrainian(2026)}]{braun2026beyond}
Braun, J.; Eickhoff, C.; and Bahrainian, S.~A. 2026.
\newblock Beyond Multiple Choice: Evaluating Steering Vectors for
  Summarization.
\newblock In \emph{Findings of the Association for Computational Linguistics:
  EACL 2026}, 3849--3884.

\bibitem[{Cao et~al.(2024)Cao, Zhang, Cao, Yin, Lin, Ma, and
  Chen}]{cao2024bipo}
Cao, Y.; Zhang, T.; Cao, B.; Yin, Z.; Lin, L.; Ma, F.; and Chen, J. 2024.
\newblock Personalized Steering of Large Language Models: Versatile Steering
  Vectors Through Bi-directional Preference Optimization.
\newblock In \emph{Advances in Neural Information Processing Systems},
  volume~37.

\bibitem[{Chawla et~al.(2021)Chawla, Ramirez, Clever, Lucas, May, and
  Gratch}]{chawla2021casino}
Chawla, K.; Ramirez, J.; Clever, R.; Lucas, G.; May, J.; and Gratch, J. 2021.
\newblock {CaSiNo}: A Corpus of Campsite Negotiation Dialogues for Automatic
  Negotiation Systems.
\newblock In \emph{Proceedings of the 2021 Conference of the North American
  Chapter of the Association for Computational Linguistics: Human Language
  Technologies}, 3167--3185.

\bibitem[{Chen et~al.(2025{\natexlab{a}})Chen, Arditi, Sleight, Evans, and
  Lindsey}]{chen2025persona}
Chen, R.; Arditi, A.; Sleight, H.; Evans, O.; and Lindsey, J.
  2025{\natexlab{a}}.
\newblock Persona Vectors: Monitoring and Controlling Character Traits in
  Language Models.
\newblock arXiv:2507.21509.

\bibitem[{Chen et~al.(2025{\natexlab{b}})Chen, Lalor, Yang, and
  Abbasi}]{chen2025personatwin}
Chen, S.; Lalor, J.~P.; Yang, Y.; and Abbasi, A. 2025{\natexlab{b}}.
\newblock {PersonaTwin}: A Multi-Tier Prompt Conditioning Framework for
  Generating and Evaluating Personalized Digital Twins.
\newblock In \emph{Proceedings of the Fourth Workshop on Generation, Evaluation
  and Metrics ({GEM})}, 774--788.

\bibitem[{Eastwick and Hunt(2014)}]{eastwick2014relational}
Eastwick, P.~W.; and Hunt, L.~L. 2014.
\newblock Relational Mate Value: Consensus and Uniqueness in Romantic
  Evaluations.
\newblock \emph{Journal of Personality and Social Psychology}, 106(5):
  728--751.

\bibitem[{Fan et~al.(2026)Fan, Murugan, Krishnan, and
  Nagaraj}]{fan2026steering}
Fan, J.~G.; Murugan, A.; Krishnan, S.; and Nagaraj, A. 2026.
\newblock Interpreting and Steering {LLM} Agents for Social Simulations.
\newblock arXiv:2609.16436.

\bibitem[{Fidone, Passaro, and Guidotti(2026)}]{fidone2026cosmos}
Fidone, G.; Passaro, L.; and Guidotti, R. 2026.
\newblock Evaluating Online Moderation via {LLM}-Powered Counterfactual
  Simulations.
\newblock In \emph{Proceedings of the AAAI Conference on Artificial
  Intelligence (AAAI-26)}.

\bibitem[{Finkel et~al.(2012)Finkel, Eastwick, Karney, Reis, and
  Sprecher}]{finkel2012online}
Finkel, E.~J.; Eastwick, P.~W.; Karney, B.~R.; Reis, H.~T.; and Sprecher, S.
  2012.
\newblock Online Dating: A Critical Analysis From the Perspective of
  Psychological Science.
\newblock \emph{Psychological Science in the Public Interest}, 13(1): 3--66.

\bibitem[{Goldberg(1992)}]{goldberg1992markers}
Goldberg, L.~R. 1992.
\newblock The Development of Markers for the {Big-Five} Factor Structure.
\newblock \emph{Psychological Assessment}, 4(1): 26--42.

\bibitem[{Hashemi et~al.(2024)Hashemi, Eisner, Rosset, Van~Durme, and
  Kedzie}]{hashemi2024rubric}
Hashemi, H.; Eisner, J.; Rosset, C.; Van~Durme, B.; and Kedzie, C. 2024.
\newblock {LLM-Rubric}: A Multidimensional, Calibrated Approach to Automated
  Evaluation of Natural Language Texts.
\newblock In \emph{Proceedings of the 62nd Annual Meeting of the Association
  for Computational Linguistics}, 13806--13834.

\bibitem[{Hu et~al.(2022)Hu, Shen, Wallis, Allen-Zhu, Li, Wang, Wang, and
  Chen}]{hu2022lora}
Hu, E.~J.; Shen, Y.; Wallis, P.; Allen-Zhu, Z.; Li, Y.; Wang, S.; Wang, L.; and
  Chen, W. 2022.
\newblock {LoRA}: Low-Rank Adaptation of Large Language Models.
\newblock In \emph{International Conference on Learning Representations}.

\bibitem[{Hu et~al.(2026)Hu, Baumann, Lupo, Collier, Hovy, and
  R{\"o}ttger}]{hu2026simbench}
Hu, T.; Baumann, J.; Lupo, L.; Collier, N.; Hovy, D.; and R{\"o}ttger, P. 2026.
\newblock {SimBench}: Benchmarking the Ability of Large Language Models to
  Simulate Human Behaviors.
\newblock In \emph{International Conference on Learning Representations}.

\bibitem[{Hu and Collier(2024)}]{hu2024personaeffect}
Hu, T.; and Collier, N. 2024.
\newblock Quantifying the Persona Effect in {LLM} Simulations.
\newblock In \emph{Proceedings of the 62nd Annual Meeting of the Association
  for Computational Linguistics}, 10289--10307.

\bibitem[{Hullman et~al.(2026)Hullman, Broska, Sun, and
  Shaw}]{hullman2026validating}
Hullman, J.; Broska, D.; Sun, H.; and Shaw, A. 2026.
\newblock This Human Study Did Not Involve Human Subjects: Validating {LLM}
  Simulations as Behavioral Evidence.
\newblock arXiv:2602.15785.

\bibitem[{Joel, Eastwick, and Finkel(2017)}]{joel2017desire}
Joel, S.; Eastwick, P.~W.; and Finkel, E.~J. 2017.
\newblock Is Romantic Desire Predictable? Machine Learning Applied to Initial
  Romantic Attraction.
\newblock \emph{Psychological Science}, 28(10): 1478--1489.

\bibitem[{Krabbe and Shi(2026)}]{krabbe2026personality}
Krabbe, T.; and Shi, X. 2026.
\newblock Do Personality-Tuned {LLMs} Make Better Social Agents?
\newblock arXiv:2609.21857.

\bibitem[{Kumar et~al.(2025)Kumar, Yan, Perepa, Yue, and
  Yao}]{kumar2025reversed}
Kumar, S. A.~S.; Yan, H.; Perepa, S.; Yue, M.; and Yao, Z. 2025.
\newblock Can {LLMs} Simulate Personas with Reversed Performance? A Systematic
  Investigation for Counterfactual Instruction Following in Math Reasoning
  Context.
\newblock Social Simulation with LLMs Workshop at COLM 2025. arXiv:2504.06460.

\bibitem[{Matz et~al.(2024)Matz, Peters, Cerf, Grunenberg, Eastwick, Back, and
  Finkel}]{matz2024attraction}
Matz, S.~C.; Peters, H.; Cerf, M.; Grunenberg, E.; Eastwick, P.~W.; Back,
  M.~D.; and Finkel, E.~J. 2024.
\newblock Can Large Language Models Detect Verbal Indicators of Romantic
  Attraction?
\newblock arXiv:2407.10989.

\bibitem[{Park et~al.(2023)Park, O'Brien, Cai, Morris, Liang, and
  Bernstein}]{park2023generative}
Park, J.~S.; O'Brien, J.~C.; Cai, C.~J.; Morris, M.~R.; Liang, P.; and
  Bernstein, M.~S. 2023.
\newblock Generative Agents: Interactive Simulacra of Human Behavior.
\newblock In \emph{Proceedings of the 36th Annual ACM Symposium on User
  Interface Software and Technology}, 1--22.

\bibitem[{Park et~al.(2024)Park, Zou, Kamphorst, Egan, Shaw, Hill, Cai, Morris,
  Liang, Willer, and Bernstein}]{park2024agents}
Park, J.~S.; Zou, C.~Q.; Kamphorst, J.; Egan, N.; Shaw, A.; Hill, B.~M.; Cai,
  C.; Morris, M.~R.; Liang, P.; Willer, R.; and Bernstein, M.~S. 2024.
\newblock {AI} Simulations Grounded in Rich Self-Reports Increase Accuracy and
  Reduce Distortion.
\newblock arXiv:2411.10109v4.

\bibitem[{Peng et~al.(2025)Peng, Gui, Brucks, Merlau
  et~al.}]{peng2025megastudy}
Peng, T.; Gui, G.; Brucks, M.; Merlau, D.~J.; et~al. 2025.
\newblock Digital Twins are Funhouse Mirrors: Five Systematic Distortions.
\newblock arXiv:2509.19088v5.

\bibitem[{Piao et~al.(2025)Piao, Yan, Zhang, Li et~al.}]{piao2025agentsociety}
Piao, J.; Yan, Y.; Zhang, J.; Li, N.; et~al. 2025.
\newblock {AgentSociety}: Large-Scale Simulation of {LLM}-Driven Generative
  Agents Advances Understanding of Human Behaviors and Society.
\newblock arXiv:2502.08691.

\bibitem[{Reece et~al.(2023)Reece, Cooney, Bull et~al.}]{reece2023candor}
Reece, A.; Cooney, G.; Bull, P.; et~al. 2023.
\newblock The {CANDOR} Corpus: Insights from a Large Multimodal Dataset of
  Naturalistic Conversation.
\newblock \emph{Science Advances}, 9(13): eadf3197.

\bibitem[{Rimsky et~al.(2024)Rimsky, Gabrieli, Schulz, Tong, Hubinger, and
  Turner}]{rimsky2024caa}
Rimsky, N.; Gabrieli, N.; Schulz, J.; Tong, M.; Hubinger, E.; and Turner, A.
  2024.
\newblock Steering {Llama} 2 via Contrastive Activation Addition.
\newblock In \emph{Proceedings of the 62nd Annual Meeting of the Association
  for Computational Linguistics}, 15504--15522.

\bibitem[{Song et~al.(2026)Song, Parwani, Matlin, Anand, Theerthala,
  Chatterjee, Zang, Kostylew, Shavit, Krier, and Riedl}]{song2026rolesteering}
Song, I.; Parwani, M.~R.; Matlin, G.; Anand, E.; Theerthala, A.; Chatterjee,
  A.; Zang, A. W.-M.; Kostylew, M.; Shavit, Y.~G.; Krier, S.; and Riedl, M.
  2026.
\newblock Role Steering of Language Models for Social Simulations.
\newblock Social Simulation with LLMs Workshop at COLM 2026. arXiv:2608.00023.

\bibitem[{Toubia et~al.(2025)Toubia, Gui, Peng, Merlau, Li, and
  Chen}]{toubia2025twin}
Toubia, O.; Gui, G.~Z.; Peng, T.; Merlau, D.~J.; Li, A.; and Chen, H. 2025.
\newblock Twin-2{K}-500: A Dataset for Building Digital Twins of over 2,000
  People Based on Their Answers to over 500 Questions.
\newblock arXiv:2505.17479.

\bibitem[{Wang, Morgenstern, and Dickerson(2025)}]{wang2025misportray}
Wang, A.; Morgenstern, J.; and Dickerson, J.~P. 2025.
\newblock Large Language Models that Replace Human Participants Can Harmfully
  Misportray and Flatten Identity Groups.
\newblock \emph{Nature Machine Intelligence}, 7(3): 400--411.

\bibitem[{Ye et~al.(2025)Ye, Chen, Wang et~al.}]{ye2025cognipair}
Ye, W.; Chen, S.; Wang, Y.; et~al. 2025.
\newblock {CogniPair}: From {LLM} Chatbots to Conscious {AI} Agents --
  {GNWT}-Based Multi-Agent Digital Twins for Social Pairing -- Dating \& Hiring
  Applications.
\newblock arXiv:2506.03543.

\bibitem[{Ye et~al.(2026)Ye, Chen, Wang et~al.}]{ye2026cognibit}
Ye, W.; Chen, S.; Wang, Y.; et~al. 2026.
\newblock Cognibit: From Digital Exhaustion to Real-World Connection Through
  Gamified Territory Control and {LLM}-Powered Twin Networking.
\newblock arXiv:2604.04351.

\bibitem[{Zhou et~al.(2024)Zhou, Zhu, Mathur, Zhang, Yu, Qi, Morency, Bisk,
  Fried, Neubig, and Sap}]{zhou2024sotopia}
Zhou, X.; Zhu, H.; Mathur, L.; Zhang, R.; Yu, H.; Qi, Z.; Morency, L.-P.; Bisk,
  Y.; Fried, D.; Neubig, G.; and Sap, M. 2024.
\newblock {SOTOPIA}: Interactive Evaluation for Social Intelligence in Language
  Agents.
\newblock In \emph{International Conference on Learning Representations}.

\bibitem[{Zhou et~al.(2026)Zhou, Wang, Ai, Wu, and Liu}]{zhou2026prosim}
Zhou, Y.; Wang, H.; Ai, Q.; Wu, Z.; and Liu, Y. 2026.
\newblock Investigating Prosocial Behavior Theory in {LLM} Agents under
  Policy-Induced Inequities.
\newblock In \emph{Proceedings of the AAAI Conference on Artificial
  Intelligence (AAAI-26)}.

\bibitem[{Z{\"u}fle et~al.(2026)Z{\"u}fle, Schmidtov{\'a}, Zouhar
  et~al.}]{zufle2026judges}
Z{\"u}fle, M.; Schmidtov{\'a}, P.; Zouhar, V.; et~al. 2026.
\newblock Calibrating {LLM} Judges for Human and {AI} Conversations.
\newblock arXiv:2609.29431.

\end{thebibliography}

\end{document}